\documentclass[lettersize,journal]{IEEEtran}
\usepackage{amsmath}
\usepackage{subeqnarray}
\usepackage{cases}
\usepackage{empheq}
\usepackage{enumerate}
\usepackage{makecell}
\usepackage[english]{babel}
\usepackage{color}
\usepackage{multirow}
\usepackage{cite}
\usepackage{graphicx} 
\usepackage{latexsym}
\usepackage{epstopdf}
\usepackage{amsfonts,amssymb}             
\usepackage{algorithm}  
\usepackage{algorithmic}
\usepackage{xcolor} 
\usepackage{stfloats}
\usepackage{amsthm}
\theoremstyle{remark}

\newtheorem{remark}{Remark}

\begin{document}
\title{Brain-Inspired Spike Echo State Network Dynamics for Aero-engine Intelligent Fault Prediction}

\author{Mo-Ran Liu, Tao Sun,  and Xi-Ming~Sun,~\IEEEmembership{Senior Member,~IEEE}
\thanks{This work was supported in part by the National Key R${\&}$D Program of China under Grant 2018YFB1700102,  in part by the National Natural Science Foundation of China under Grants 61890920 and 61890921, and in part by the the China Postdoctoral Science Foundation funded project under Grant 2022TQ0179. (Corresponding author: T. Sun.)}
\thanks{M.-R. Liu and X.-M.~Sun are with the Key Laboratory of Intelligent Control and Optimization for Industrial Equipment of Ministry of Education,  Dalian University of Technology, Dalian 116024, China (e-mail: liumoran@mail.dlut.edu.cn; sunxm@dlut.edu.cn).}
\thanks{T. Sun is with the Department of Automation, Tsinghua University, Beijing 100084, China(e-mail: tao.sun.meng@gmail.com).}}

\markboth{IEEE TRANSACTIONS ON INSTRUMENTATION AND MEASUREMENT}%
{Liu \MakeLowercase{\textit{et al.}}: Brain-Inspired Spike Echo State Network Dynamics for Aero-engine Intelligent Fault Prediction}

\maketitle

\begin{abstract}
Aero-engine fault prediction aims to accurately predict the development trend of the future state of aero-engines, so as to diagnose faults in advance. Traditional aero-engine parameter prediction methods mainly use the nonlinear mapping relationship of time series data but generally ignore the adequate spatio-temporal features contained in aero-engine data. To this end, we propose a brain-inspired spike echo state network (Spike-ESN) model for aero-engine intelligent fault prediction, which is used to effectively capture the evolution process of aero-engine time series data in the framework of spatio-temporal dynamics. In the proposed approach, we design a data spike input layer based on Poisson distribution inspired by the spike neural encoding mechanism of biological neurons, which can extract the useful temporal characteristics in aero-engine sequence data. Then, the temporal characteristics are input into a spike reservoir through the current calculation method of spike accumulation in neurons, which projects the data into a high-dimensional sparse space. In addition, we use the ridge regression method to read out the internal state of the spike reservoir. Finally, the experimental results of aero-engine states prediction demonstrate the superiority and potential of the proposed method.
\end{abstract}


\IEEEpeerreviewmaketitle

\section{Introduction}
\noindent \IEEEPARstart{A}{s} the core power source of civil or military aircrafts, aero-engine is a highly complex and precise pneumatic thermal mechanical system, and its working state has a direct or indirect impact on the safety, reliability, economy and other performance problems of aircrafts \cite{tahan2017performance,sun2020adaptive,shi2020bumpless,wen2022practical,sun2021optimal}. Since the aero-engine works in harsh environments of high temperature, high pressure and strong vibration for a long time, the probability of failure increases accordingly, which affects the working performance of the aero-engine and even causes bad flight accidents. In fact, the surge fault is one of the most representative and destructive faults of aero-engines or aircrafts. When the aircraft flies under extreme conditions, the aero-engine system is affected by the external environment \cite{aiswarya2018efficient}, which makes the output pressure of the compressor less than the downstream of the system and the backward flow of high-pressure gas.
\begin{figure}[ht]
	\centering
	\includegraphics[height=0.75\textwidth,width=1.\hsize]{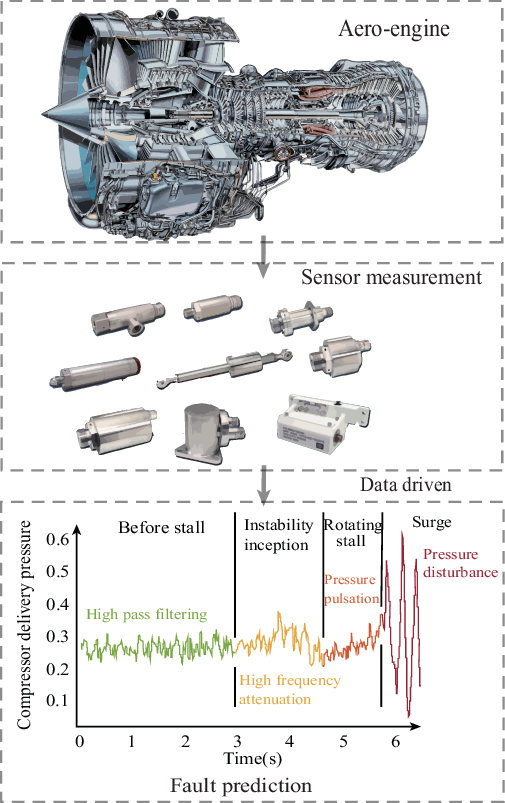}
	\caption{{Schematic diagram of data driven fault diagnosis of an aero-engine system.}}\label{fig11}
\end{figure}
Accordingly, the aero-engine produces violent vibration and overtemperature at the hot end, which causes the aircraft to shutdown in the air. The research shows that aero-engine fault occurrence generally includes instability inception, rotating stall and surge \cite{khaleghi2015stall,pan2021effect}, where rotating stall is the precursor of surge (i.e., as shown in Figure \ref{fig11}). Since the aero-engine data has strong nonlinear and high-dimensional characteristics in actual operation, the fault diagnosis of an aero-engine generally relies on the empirical criteria of the crew and ground researchers. However, due to the subjectivity of users, it is often impossible to accurately predict faults using empirical estimation methods, so that the aircraft operation fault cannot be judged and dealt with in time. Therefore, it is of great practical significance for the safe, stable and efficient operation of an aircraft to scientifically predict the potential fault factors of aero-engines and terminate the surge fault in the early stage.

The aero-engine fault diagnosis method mainly includes the method based on the mechanism model and the method based on the data analysis. In the early stage, the forecasting method based on the mechanism model was the main research idea of aero-engine fault diagnosis\cite{chen2021bilateral}. For example, paper \cite{urban1973gas} proposed a fault influence coefficient matrix method for aero-engine physical model, which is used in many traditional fault diagnosis systems. By using a fault map model, each state of the aero-engine in \cite{davison2001development} is mapped to a point or a region, where researchers can use fault maps to qualitatively delineate areas and identify fault types according to empirical criteria. In addition, the diagnostic methods based on nonlinear Kalman filter algorithm are designed for the Gaussian noise environment and nonlinear properties of aero-engine parameters. For instance, a fuzzy adaptive unscented Kalman filter algorithm is developed in \cite{han2018study}. In the presence of gas path measurement uncertainty, paper \cite{lu2019novel} improved a Kalman filter by using a multi-step recursive estimation strategy and self-tuning buffers. According to the above fault diagnosis report based on the mechanism model, the real aero-engine is an extremely complex nonlinear system, so the fault diagnosis method based on mechanism model has high requirements on the design accuracy of the model.

The fault diagnosis methods based on data analysis may not require precise physical models but sufficient judgmental experience and historical data. So far, data-driven based intelligent fault diagnosis methods mainly include artificial neural network  \cite{ren2022aero,li2021tiny,an2022interpretable,jia2022multiscale}, support vector machine\cite{sun2021new}, extreme learning machine \cite{zhao2020multiple}, etc. In particular, an extended least squares support vector machine in \cite{zhao2020extended} is presented to measure the failure of aero-engines. Moreover, the extreme learning machine based on quantum behavioral particle swarm optimization is applied to the diagnosis of gas turbofan engines in \cite{yang2016aero}. In fact, the artificial neural networks have more advantages than above machine learning methods in the area of nonlinear prediction \cite{qiu2018rgb,qiu2019outdoor}. For example, paper \cite{liu2020research} utilized a fault diagnosis method based on the back propagation neural network for the failure of the aero-engine sensors, while the back propagation neural network method cannot memorize the past sequence data. Instead, paper \cite{yuan2016fault} designed a long short-term memory neural network for fault diagnosis in the cases of complex environment \cite{yanhua2022adaptive}. However, the long short-term memory neural network can hardly extract the temporal features of the data, and the solution of its weight parameters requires an additional optimization algorithm.

Therefore, the above observations motivate the following problems naturally: 
1) How to provide a data-driven intelligent diagnosis method to solve the fault prediction problem of aero-engines? 
2) How to introduce a spike neural mechanism with brain-inspired learning to deeply extract the spatio-temporal feature of an aero-engine time series with multi-dimensional, multi-scaled and multi-modal characteristics?
3) How to preserve bio-interpretability, sparsity and memory of a neural network
model for aero-engines? Inspired by the above problems, since aero-engines with highly coupled subsystems are sensitive to complex working environments, the research on aero-engine fault prediction in the area of aero-engines is also an inevitable and interesting topic. As far as we know, adequately capturing spatio-temporal feature is always one of the most challenging topics in the field of fault diagnosis and data prediction, and there is not report on the aero-engine fault prediction of the so-called “brain-inspired spike echo state network” from the viewpoint of spatio-temporal features, sparsity and memory. This paper is motivated by the above discussions.

Overall, the main contributions of this work are summarized as follows:
\begin{itemize}
	\item[$\bullet$] Inspired by the spike encoding mechanism in biological neurons, we propose a spike input layer that utilizes Poisson distribution to randomly spike encode the aero-engine time series data, which emphasizes a temporal dynamic characteristic.
	\item[$\bullet$] In order to further extract the spatial dynamic features of the aero-engine data, we designed a spike reservoir that uses a current calculation method of spike accumulation in neurons, which can project the temporal characteristics of the spike sequence into a high-dimensional space.
	\item[$\bullet$] A brain-inspired spike echo state network (Spike-ESN) model is designed for aero-engine intelligent fault prediction. Compared with the echo state network (ESN) model \cite{zhao2020aero} and the traditional auto-regressive and moving average (ARMA) model \cite{LIU2011115}, the proposed Spike-ESN model with both sparsity and memory can effectively capture the evolution process of aero-engine time series data in spatio-temporal space, which plays a key role in real-time prediction of aero-engine states. In addition, we have conducted a lot of experiments on the actual aero-engine datasets with faults to prove the effectiveness of Spike-ESN for aero-engine fault prediction.
\end{itemize}

The remainder of this article is organized as follows. Section II introduces the time series prediction problem of aero-engine fault diagnosis. Section III designs a brain-inspired spike echo state network with the spike neural mechanism. Some numerical simulation results are checked in Section IV. Main conclusions of the paper are given in Section V.\\

\section{Problem preliminaries}
\noindent According to the statistics of China in the early decade of the 21st century, 60$\%$ of aircraft mechanical failure accidents are caused by the aero-engine fault, and more than 30$\%$ of the daily maintenance costs of aircraft are generated by aero-engines. In order to ensure the safety and stability of aero-engines, it is indispensable to monitor the fault and predict the life of aero-engines. Generally, scientists predict failures through various methods of detecting and calculating trends in the operating state of an aero-engine system.
\begin{figure}[htbp]
	\centering
	\includegraphics[height=0.6\textwidth,width=1.\hsize]{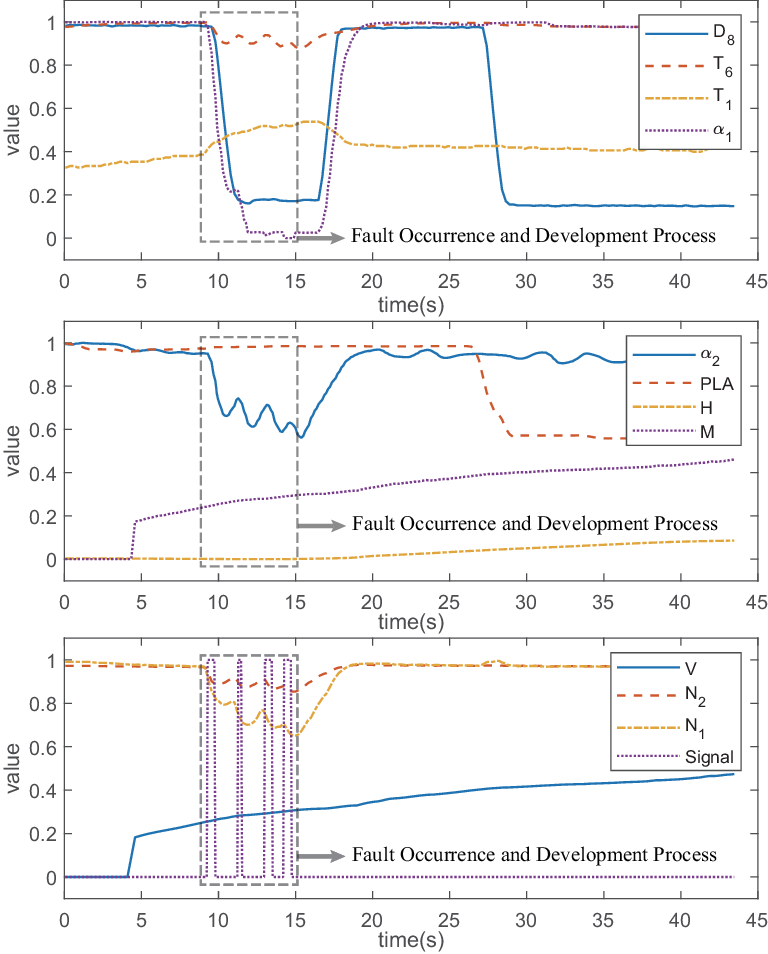}
	\caption{{Schematic diagram of operating data of an aero-engine.}}\label{fig3}
\end{figure}
In the actual situation, the fault diagnosis system judges the abnormal signals sent out before the aero-engine failure by studying the characteristics of the system, so that the crew can take crucial measures for the corresponding signals to prevent the occurrence of faults.

For example, the aircraft data with faults from an Aero-engine Research Institute is shown in Figure \ref{fig3}, where ${{D}_{8}}$ is the ambient pressure, ${{T}_{6}}$ and ${{T}_{1}}$ are the exhaust temperature and the inlet temperature respectively, ${{\alpha }_{1}}$ and ${{\alpha }_{2}}$ are the compressor opening angles, $PLA$ is the power lever angle, $H$ is the flight height, $M$ represents the Mach number, $V$ represents the flight speed, ${{N}_{2}}$ and ${{N}_{1}}$ are high and low pressure rotor speeds respectively, $signal$ indicates the occurrence of a fault. As can be seen from Figure \ref{fig3}, around the 200th data point, the ambient pressure ${{D}_{8}}$ droped sharply. At the same time, the high-pressure rotor speed ${{N}_{2}}$ and the low-pressure rotor speed ${{N}_{1}}$ decreased significantly, while $signal$ indicated that the aero-engine has a fault. Based on this, it is feasible to predict aircraft engine failures through data trends.

In fact, the fault prediction of aero-engines can be regarded as a time series prediction problem. Empirically, it is assumed that the statistical evaluation index data is represented as a set of univariate time series $\left\{ {{z}_{i}}\in {{\mathbb{R}}} \right\}_{i=1}^{N}$, where ${{z}_{i}}$ is the ${{i}^{th}}$ point element. Then, when the prediction time step is $\tau $, the time series data ${{z}_{i}}$ can be expressed as follows,
\begin{eqnarray}\label{EQ15}
	{{z}_{i}}=f({{z}_{i-\tau }},{{c}_{i-\tau }},\varphi ),
\end{eqnarray}
where ${{c}_{i-\tau }}$ is a covariate that jointly affects the prediction results with ${{z}_{i-\tau}}$, $f(\cdot )$ represents the nonlinear mapping relationship, $\varphi$ is a random parameter representing the random fluctuation of the time series.\\

\begin{figure*}[ht]
	\centering
	\includegraphics[height=0.32\textwidth,width=1\hsize]{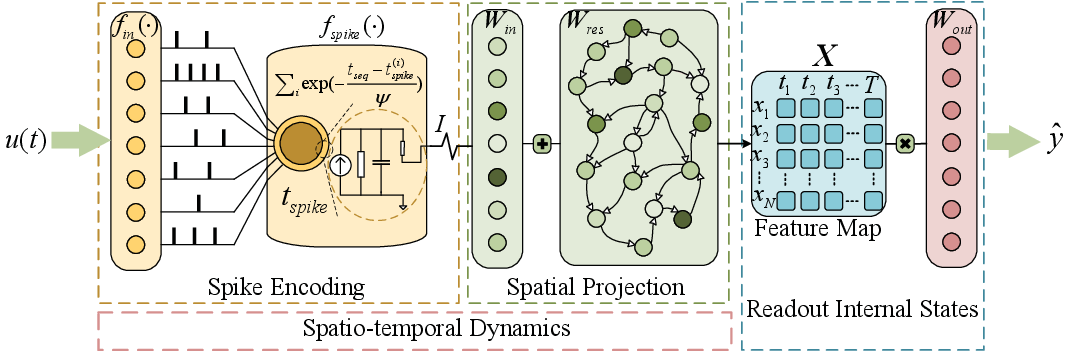}
	\caption{{Spike echo state network. It consists of a spike input layer, a spike reservoir and an output layer.}}\label{fig1}
\end{figure*}

\section{Methodology: spike echo state network}
\noindent In this section, in order to solve the above time series prediction problem of aero-engine fault diagnosis, we establish a spike echo state network (Spike-ESN) model composed of spike input layer, spike reservoir and output layer for aero-engine fault prediction. 

\subsection{Spike input layer}
\noindent The spike input layer can encode the input signal into a spike sequence, which emphasizes the temporal features of data in order to extract critical information.

First, we treat the normalized input data $u(t)\in \mathbb{R}$ as the probability of the spike interval, which determines the average of the spike interval, as follows,
\begin{eqnarray}\label{EQ3}
	{{h}_{\kappa }}(t)={{N}_{sam}}\times \frac{{{U}_{max}}-u(t)}{{{U}_{max}}-{{U}_{min}}},
\end{eqnarray}
where ${{h}_{\kappa }}(t)\in \mathbb{R}$ represents the average interval of spike generated by the input data $u(t)$, ${{U}_{max}}$ and ${{U}_{min}}$ are the maximum and minimum values of the input time series data, respectively. ${{N}_{sam}}\in {{\mathbb{Z}}^{+}}$ is the spike sampling times of the spiking neuron. With the increase of spikes sampling times, the input data are projected to a higher time dimension, which is better for extracting effective features.

Then, we use the following Poisson distribution to generate the spike sequence,
\begin{eqnarray}\label{EQ2}
	P(k)=\frac{{{m}^{k}}}{k!}{{e}^{-m}},
\end{eqnarray}
where $k$ represents the number of events, $m$ is the mean and variance of Poisson distribution.
\begin{remark}
	Poisson encoding is one of the effective methods for spike input layers. In theory, The Poisson distribution is suitable for describing the number of random events occurring per unit time, which corresponds exactly to the definition of spike issuance rate. Moreover, Poisson encoding can transform the input image or signal into a spike sequence that conforms to the Poisson process, so that the continuity and sparsity of the input information can be maintained.
\end{remark}

Next, all intervals $K (t)$ can be randomly generated with ${{h}_{\kappa }}(t)$ as the mean of the Poisson distribution shown in formula \eqref{EQ2}, i.e.,
\begin{eqnarray}\label{EQ16}
	K (t)={{\left[ {{\kappa }_{1}}(t),{{\kappa }_{2}}(t),\cdots ,{{\kappa }_{{{N}_{int}}}}(t) \right]}^{T}},
\end{eqnarray}
where for any $l\in \left\{ 1,2,\cdots ,{{N}_{int}} \right\}$, ${{\kappa }_{l}}(t)\in \left\{ 1,2,\cdots \\ ,{{N}_{sam}} \right\}$, ${{N}_{int}}\in {{\mathbb{Z}}^{+}}$ is the number of spike intervals, ${{N}_{sam}}$ and ${{N}_{int}}$ satisfy the relationship as follows,
\begin{eqnarray}\label{EQ17}
	\sum\limits_{l=1}^{{{N}_{\operatorname{int}}}}{{{\kappa }_{l}}(t)}\le {{N}_{sam}}.
\end{eqnarray}

Finally, the spike sequences are generated by intervals, i.e., according to the following rules,
\begin{eqnarray}\label{EQ4}
	{{s}_{i}}(t)=\left\{ \begin{array}{*{35}{l}}
		1\text{,  \qquad    }i=\sum\limits_{j=1}^{k}{{{\kappa }_{j}}(t)}  \\
		0\text{, \qquad    otherwise   }  \\
	\end{array} \right.,
\end{eqnarray}
where $i\in \left\{ 1,2,\cdots {{N}_{sam}} \right\}$ is the position of the element in the spike sequence, $k\in \left\{ 1,2,\cdots ,{{N}_{int}} \right\}$, $j\in {{\mathbb{Z}}^{+}}$.

Based on the above method, the spike input layer ${{f}_{in}}(\cdot )$ can convert the input signal into a spike sequence as follows,
\begin{eqnarray}\label{EQ1}
	{{f}_{in}}(u(t))={{[{{s}_{1}}(t),{{s}_{2}}(t),...,{{s}_{{{N}_{sam}}}}(t)]}^{T}}\in {{\left\{ 0,1 \right\}}^{{{N}_{sam}}}},
\end{eqnarray}
where ${[{{s}_{1}}(t),{{s}_{2}}(t),...,{{s}_{{{N}_{sam}}}}(t)]}^{T}$ is the spike sequence converted from the input data, for any $i\in \left\{ 1,2,\cdots {{N}_{sam}} \right\}$, ${{s}_{i}}(t)\in \{0,1\}$, where 1 and 0 represent the activation and inhibition of the spike neuron, respectively. 

\subsection{Spike reservoir}
\noindent The spike reservoir is a sparse network composed of large-scale neurons randomly connected, which can memorize information by adjusting the internal weights of the network. 

First, we generate spike reservoirs in order to represent the input data in a high-dimensional and non-linear manner, as follows,
\begin{eqnarray}\label{EQ5}
	{{\boldsymbol W}_{res}}=\rho \cdot \frac{\boldsymbol W}{{{\lambda }_{\max }}(\boldsymbol W)}\in {{\mathbb{R}}^{{{N}_{res}}\times {{N}_{res}}}},
\end{eqnarray}
where ${{\boldsymbol W}_{res}}$ is the random fixed internal weight of the reservoir, $\boldsymbol W\in {{\mathbb{R}}^{{{N}_{res}}\times {{N}_{res}}}}$ is a matrix with sparse degree $\eta$ randomly generated from a uniform distribution on the interval $[-1,1]$. When the data is projected into the high-dimensional and sparse space, the readout of the state will be easier. ${{N}_{res}}$ is the number of spike neurons in the spike reservoir, which needs to be set at a moderate level. If this parameter is set too small, the internal features cannot be projected into the high-dimensional space to obtain the best results. On the contrary, if the parameter is set too large, it will cause the introduction of redundant information and difficulties in decoupling state. ${{\lambda }_{\max }} (\boldsymbol W)$ is the maximum eigenvalue of matrix $\boldsymbol W$, $\rho$ is the spectral radius that controls and determines the generation of ${{\boldsymbol W}_{res}}$, i.e., the maximum eigenvalue of the adjacency matrix $\textbf{W}_{res}$ of the reservoir is limited by $\rho$.

Then, the update mode of the internal state $x(t)\in {{\mathbb{R}}^{{{N}_{res}}}}$ of the spike reservoir can be expressed as follows,
\begin{eqnarray}\label{EQ6}
\left\{ \begin{array}{*{35}{l}}
	\displaystyle x(t)=\tanh ({{\boldsymbol {W}}_{in}}{{f}_{spike}}({{t}_{spike}})+{{\boldsymbol {W}}_{res}}x(t-1)),\vspace{1ex}\\
	\displaystyle {{t}_{spike}}={{[t_{spike}^{(1)},t_{spike}^{(2)},\cdots ,t_{spike}^{({{N}_{int}})}]}^{T}},\vspace{1ex}\\
	\displaystyle  t_{spike}^{(i)}=\text{Position}({{s}_{act}^{(i)}})\in \left\{ 1,2,\cdots {{N}_{sam}} \right\},\vspace{1ex}\\
	\displaystyle {{f}_{spike}}(\cdot )={{[I(1),I(2),\cdots ,I({{N}_{sam}})]}^{T}}\in {{\mathbb{R}}^{{{N}_{sam}}}},\vspace{1ex}\\
	\displaystyle I({{t}_{seq}})=\sum\limits_{i=1}^{{{N}_{\operatorname{int}}}}{\exp (-\frac{{{t}_{seq}}-t_{spike}^{(i)}}{\psi })}\in \mathbb{R},\\
	\displaystyle i\in \left\{ 1,2,\cdots {{N}_{int}} \right\},\\
\end{array} \right.
\end{eqnarray}
where ${{\boldsymbol W}_{in}}\in {{\mathbb{R}}^{{{N}_{res}}\times {{N}_{sam}}}}$ is the input weight matrix of the reservoir, which is generated from the uniform distribution on the interval $[-1,1]$, ${{f}_{spike}}(\cdot )$ contains the external input current $I({{t}_{seq}})$ of the neuron at each sampling time ${{t}_{seq}}=\left\{ 1,2,\cdots ,{{N}_{sam}} \right\}$, $t_{spike}^{(i)}$ is the occurrence time of the ${{i}^{th}}$ spike, $\psi $ is the time constants of synaptic currents, which controls the overall magnitude level of the current emitted.
${{s}_{act}^{(i)}}$ is the $i^{th}$ element with the value of 1 in the spike sequence ${[{{s}_{1}}(t),{{s}_{2}}(t),...,{{s}_{{{N}_{sam}}}}(t)]}^{T}$, i.e., the spike neuron is activated at this moment, and the function $\text{Position}({{s}_{act}^{(i)}})$ represents the specific position of the element ${{s}_{act}^{(i)}}$ in the spike sequence. In particular, when $t=1$, the reservoir is initialized in the empty state, i.e., $x(0)=0$.
\begin{remark}
According to (9), the internal state $x(t)$ is converged when the maximum eigenvalue of $\textbf{W}_{res}$ is less than 1, i.e., the historical input is gradually forgotten as the network is updated. In short-term time series prediction, the future data are only related to the recent historical data.
\end{remark}

Finally, after the reservoir is updated periodically, all internal states $x(t)$ are integrated in the state collection matrix $\boldsymbol X$, as follows,
\begin{eqnarray}\label{EQ8}
	\boldsymbol X=[x(1),x(2),...,x(T)]\in {{\mathbb{R}}^{{{N}_{res}}\times T}},
\end{eqnarray}
where $[\cdot \text{ , }\cdot ]$ represents the parallel connection between state vectors, and $T$ is the number of input samples. 

\subsection{Output layer}
\noindent The output layer ${{\boldsymbol W}_{out}}\in {{\mathbb{R}}^{1\times {{N}_{res}}}}$ is a weight matrix that needs to be optimized \cite{jaeger2001echo}, which can parse the internal state of the previous layer into output data, as follows,
\begin{eqnarray}\label{EQ9}
	\hat{y}={{\boldsymbol W}_{out}}\boldsymbol X,
\end{eqnarray}
where $\hat{y}=[\hat{y}(1),\hat{y}(2),...,\hat{y}(T)]\in {{\mathbb{R}}^{1\times T}}$ is the predicted data.

In order to make the model output data closer to the real value, the optimization problem of the output weight can be constructed as follows,
\begin{eqnarray}\label{EQ10}
	\underset{{{\boldsymbol W}_{out}}}{\mathop{\min }}\,\left\| \hat{y}-{{\boldsymbol W}_{out}}\boldsymbol X \right\|_{2}^{2}+\mu  {{\left\| {{\boldsymbol W}_{out}} \right\|}_{2}^{2}},
\end{eqnarray}
where ${{\left\| \cdot  \right\|}_{2}}$ represents the ${{L}_{2}}$ norm, $\mu $ is the regularization coefficient, which is used to add the penalty term to the optimization problem, so as to avoid the network from falling into overfitting. The larger the regularization coefficient, the simpler the model and the less risk of overfitting, but it may lead to underfitting. The smaller the regularization coefficient, the more complex the model and the greater the risk of overfitting, but it may improve the performance on the training set.

Generally, the above optimization problem is solved by ridge regression with pseudoinverse, as follows,

\begin{eqnarray}\label{EQ11}
	{{\hat{\boldsymbol W}}_{out}}=y{{\boldsymbol X}^{T}}{{(\boldsymbol X{{\boldsymbol X}^{T}}+\mu \boldsymbol I)}^{-1}},
\end{eqnarray}
where $\boldsymbol I$ is an identity matrix, $y$ is the target output signal.

\begin{figure}[h]
\centering
\includegraphics[height=0.55\textwidth,width=1.\hsize]{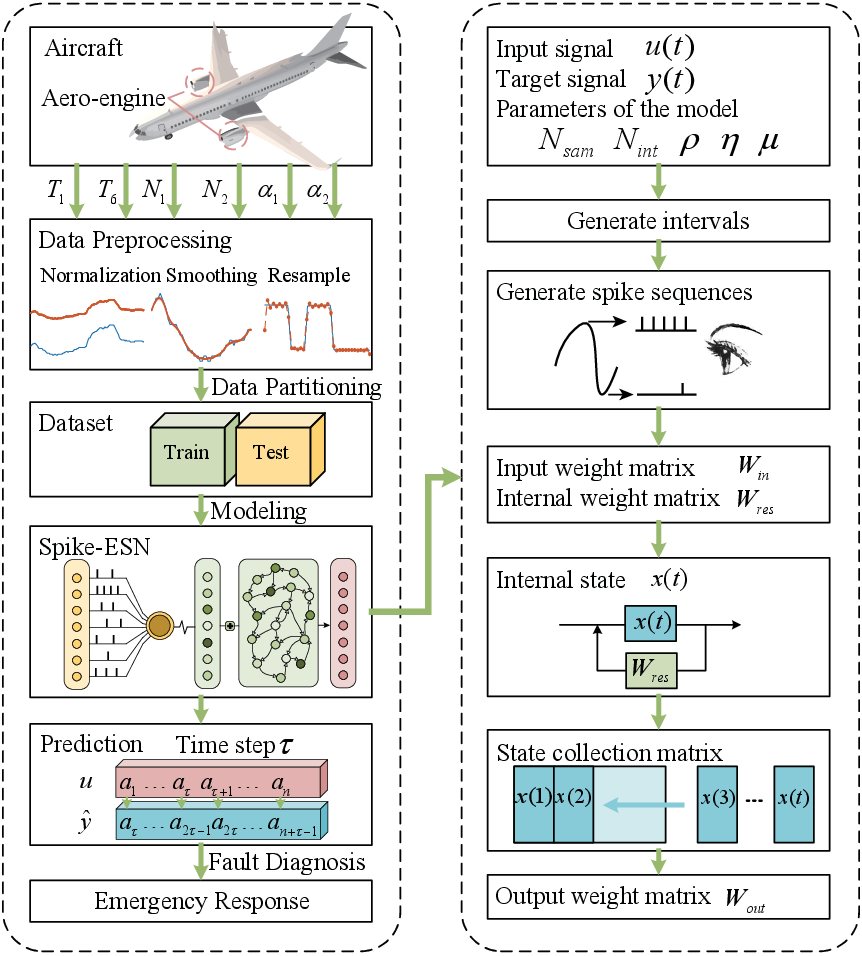}
\caption{{Process of using Spike-Esn for aircraft fault early warning.}}\label{fig13}
\end{figure}

Above all, the Spike-ESN structure is shown in Figure \ref{fig1}, and the establishment process of the Spike-ESN model is as shown in Algorithm \ref{al1}. Further, the aircraft fault warning process is shown in Figure \ref{fig13}. Since the scheme in Figure \ref{fig13} is only one of the parts of the aircraft failure prediction, it is necessary to consider the solution more completely. In practical engineering, the operating environment of aircraft is changeable, so the optimum model parameters need to be adjusted accordingly. The hyperparameter $\psi$ have an impact on the network by keeping the internal state at the right level. Since our dataset is recorded during the actual flight of the aircraft, it contains almost all the operating conditions. Therefore, our network states do not change much as the actual operating conditions change. When the input data produces a large state $x(t)$, the value of hyperparameter $\psi$ increases in fixed steps. Conversely, when the state $x(t)$ is small, the value of hyperparameter $\psi$ decreases in fixed steps. Finally, the network can start retraining until the internal state value satisfies the stop condition.
\begin{remark}
Spike-ESN consists of three parts: spike encoding, spatial projection, and readout of internal states. First, the spike encoding part strengthens the feature correlation of sequence with historical data from the perspective of high-dimensional projection in time. Then, the spatial projection divides the features into several easily distinguishable subspaces, which reduces the complexity, coupling and redundancy. Finally, the readout of the internal state part is able to solve the desired law from the adequate features exhibited by the first two parts. Since the training method of Spike-ESN is to solve the least squares loss function by regression, the feature extraction of the network depends largely on whether the features obtained from the front-end of the network are adequate and sparsely separable. Therefore, the spike encoding and spatial projection are proposed in this paper to fully display the temporal features in the high-dimensional sparse space. It is believed that a high-dimensional sparse space can help decoupling, because the sparse projection allows non-zero dense data to be clustered together and form features and patterns for subsequent network extraction. The high-dimensional sparse space can decompose the high-dimensional sparse features into different subspaces, which reduces the complexity and redundancy of the features and improves the extractability and interpretability of the features.
\end{remark}
\begin{algorithm}[htb] 
	\caption{Training Algorithm of Spike-ESN.} 
	\label{al1} 
	\begin{algorithmic}[1] 
		\REQUIRE ~~\\ 
		Input signal $u(t)$, target output signal $y(t)$, spectral radius $\rho $, sparsity $\eta $, regularization coefficient $\mu $, the spike sampling times ${{N}_{sam}}$, the number of spike intervals ${{N}_{int}}$;\\
		\STATE Calculate spike average interval ${{h}_{\kappa }}(t)\in \mathbb{R}$ by using formula \eqref{EQ3}; 
		\FOR{$i = 1$ \TO $N_{int}$}
		\STATE Generate random intervals ${{\kappa }_{i}}(t)$ in Poisson distribution by using formula \eqref{EQ2}-\eqref{EQ16}; 
		\ENDFOR
		\FOR{$i = 1$ \TO $N_{sam}$}
		\STATE Generate spike sequences ${{s}_{i}}(t)$ for each input signal by using formula \eqref{EQ4};
		\ENDFOR
		\STATE Randomly generate input weight matrix ${{\boldsymbol W}_{in}}\in {{\mathbb{R}}^{{{N}_{res}}\times {{N}_{sam}}}}$ in uniform distribution; 
		\STATE Calculate the internal weight matrix ${{\boldsymbol W}_{res}}\in {{\mathbb{R}}^{{{N}_{res}}\times {{N}_{res}}}}$ with the sparsity of $\eta$ and the spectral radius of $\rho$ by using formula \eqref{EQ5}; 
		\STATE Initialize an empty state collection matrix $\boldsymbol X$; 
		\FOR{$t = 1$ \TO $T$}
		\STATE Calculate the internal state $x(t)\in {{\mathbb{R}}^{{{N}_{res}}}}$ at each time by using formula \eqref{EQ6}, and add $x(t)$ to the state collection matrix $\boldsymbol X$;
		\ENDFOR
		\STATE Calculate the output weight matrix ${{\boldsymbol W}_{out}}$ by using formula \eqref{EQ11};
		\ENSURE ~~\\ 
		Output weight matrix ${{\boldsymbol W}_{out}}\in {{\mathbb{R}}^{1\times {{N}_{res}}}}$;
	\end{algorithmic}
\end{algorithm}

\section{Simulation examples}
\noindent In this section, we present the numerical results of Spike-ESN model for aero-engine fault prediction.

\subsection{Experimental data and parameter setting}
\noindent The data we used comes from an aero-engine research institute, which standardized the data in order to comply with the principle of confidentiality. First, we divide the dataset after preprocessing the aero-engine data. The exhaust temperature $T_6$, the combustion chamber temperature $T_1$, high-pressure rotor speed $N_2$, low-pressure rotor speed $N_1$, compressor opening angle $\alpha_1$ and $\alpha_2$ are selected as prediction targets, where a strong correlation exists between these parameters. 
\begin{figure}[h]
	\centering
	\includegraphics[height=0.35\textwidth,width=1.\hsize]{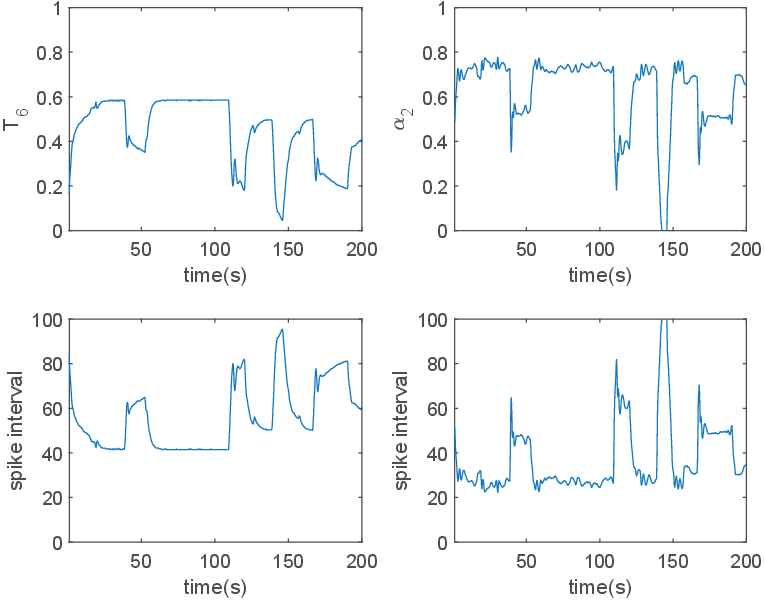}
	\caption{{The average spike interval of the exhaust temperature $T_6$ and the compressor opening angle $\alpha_2$ in the training dataset.}}\label{fig4}
\end{figure}
\begin{figure}[h]
	\centering
	\includegraphics[height=0.23\textwidth,width=1.\hsize]{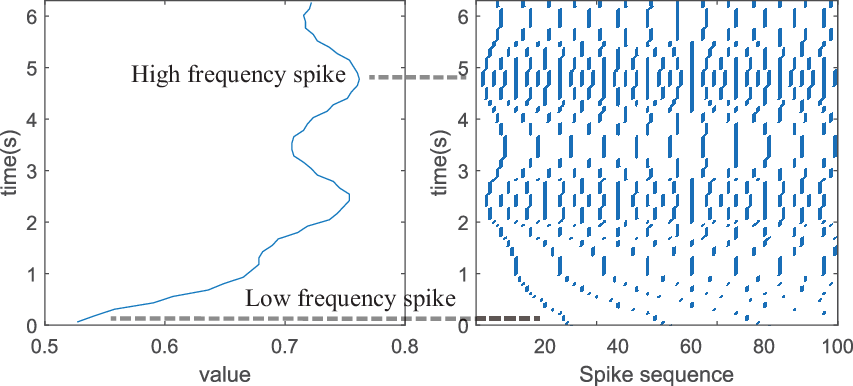}
	\caption{{Spike encoding result of a period of data of compressor opening angle $\alpha_2$.}}\label{fig5}
\end{figure}
In order to make the model learn all the features in the process of aero-engine operation as much as possible, including the normal operation and the faults caused by sudden reduction of ambient pressure $D_8$, the experiment takes a period of flight data of the aircraft as the training set, where the fault data is randomly and smoothly inserted into the training set.

Then, the dataset is spiked by using the Poisson distribution coding method in Algorithm \ref{al1}. For example, when the number of spike samples is set to 100, the average interval of the exhaust temperature $T_6$ and the compressor opening angle $\alpha_2$ in the training dataset is shown in Figure \ref{fig4}. In addition, a period of data of compressor opening angle $\alpha_2$ are spike encoded as shown in Figure \ref{fig5}, where the frequency of spike sequence generated by large value is high, and the frequency of spike sequence generated by small value is low instead.

In the experiment, we use the proposed Spike-ESN model, the echo state network (ESN) \cite{zhao2020aero}, auto regression and moving average (ARMA) \cite{LIU2011115}, convolution neural network (CNN) \cite{zhao2017convolutional}, convolution neural network (LSTM) \cite{yan2018financial} and Transformer \cite{mohammadi2020transformer} for comparison. In order to make the prediction results as fair as possible, the following parameter settings are designed: the key parameters of the Spike-ESN model and the ESN model are set the same, where the reservoir dimension ${{N}_{res}}$ is set to 100, the spectral radius $\rho $ is set to 0.9, the sparsity $\eta $ is set to 0.1, the regularization coefficient $\mu$ of regression calculation is set to ${{10}^{-8}}$. In order to make the echo state within the nonlinear range of $\tanh()$ activation function, the scaling factor is set to 0.8. For the spike neurons, the spike sequence length is set to 100, and the time constant $\psi $ of synaptic currents is set to 5000. Before the training, the echo state network needs to be initialized. The first 200 data points in the dataset are used for initialization, which are not included in the network evaluation. 

In addition, in order to verify the advantage of the proposed Spike-ESN model in aero-engine fault prediction, ARMA is set as the parameter under optimal conditions. When ARMA model is set to $ARMA(4,4)$, the effect of the model reaches the optimal.
Actually, the effect of ARMA model increases significantly with the increase of parameters, but further increase of parameters will lead to increased computation rather than better effect. Therefore, AR and MA in ARMA model are set to 4 parameters respectively, i.e., $ARMA(4,4)$. Then the structure of CNN is set as 2 COV1D layers-1 MAXPooling1D layer-1 Flatten layer-3 Dense layers, and the sliding window length is set to 15. Next, we set 4 neurons in the LSTM layer and set the number of training cycles to 100. And we set the parameters of Transformer to 4 multi-headed attention layers, 128 feature nodes, 128 fully connected neurons, 2 layers of encoder, 1 layer of decoder, and sliding window is also set to 15.
Based on the analysis above, the main settings of the 6 networks are shown in Table \ref{tab2}.
\begin{table*}[ht]
	\centering
	\renewcommand{\arraystretch}{1.2}
	\caption{{Main settings of the 6 networks (CNN, Transformer, LSTM, ARMA, ESN, Spike-ESN).}}\label{tab2}
	\begin{tabular}{ccccccc}
		\hline
		\multirow{2}{*}{Model}&\multicolumn{6}{c}{\multirow{2}{*}{Main Settings}}\\
		&&&&&&\\
		\hline
		CNN&\multicolumn{6}{c}{2 COV1D layers, 1 MAXPooling1D layer, 1 Flatten layer, 3 Dense layers}\\
		Transformer&\multicolumn{6}{c}{4 multi-headed attention layers, 128 feature nodes, 128 neurons, 2 encoders, 1 decoder}\\
		{LSTM}&\multicolumn{6}{c}{{4 neurons, 100 training cycles}}\\
		{ARMA}&\multicolumn{6}{c}{{$ARMA(4,4)$}}\\
		{ESN}&\multicolumn{6}{c}{$N_{res}=100$, $\rho=0.9$, $\eta=0.1$, $\mu = 10^{-8}$, scaling factor is 0.8}\\
		{\textbf{Spike-ESN}}&\multicolumn{6}{c}{$N_{res}=100$, $\rho=0.9$, $\eta=0.1$, $\mu = 10^{-8}$, scaling factor is 0.8, $N_{sam}=100$, $\psi=5000$}\\
		\hline
	\end{tabular}
\end{table*}
\begin{figure*}[h]
	\centering
	\includegraphics[height=0.6\textwidth,width=1.0\hsize]{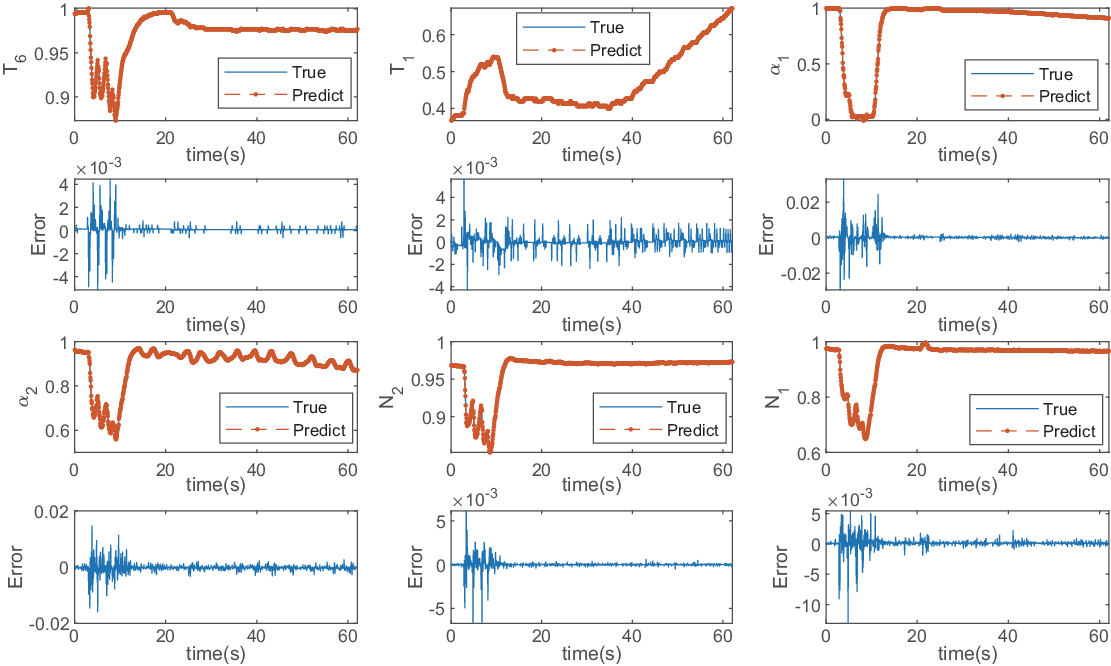}
	\caption{{Prediction results of aero-engine parameters $T_6$, $T_1$, $\alpha_{1}$, $\alpha_{2}$, $N_2$, $N_1$ when the prediction step is 1.}}\label{fig6}
\end{figure*}
Finally, Root mean square error (i.e., $RMSE=\sqrt{\sum{{{(\hat{y}-y)}^{2}}}/n}$) and mean absolute percentage error (i.e., $MAPE=\sum{\left| (\hat{y}-y)/y \right|}/n$) are used as quantitative evaluation indicators for verifying the prediction results.

\subsection{Experimental results}
\begin{figure*}[ht]
	\centering
	\includegraphics[height=0.6\textwidth,width=1.\hsize]{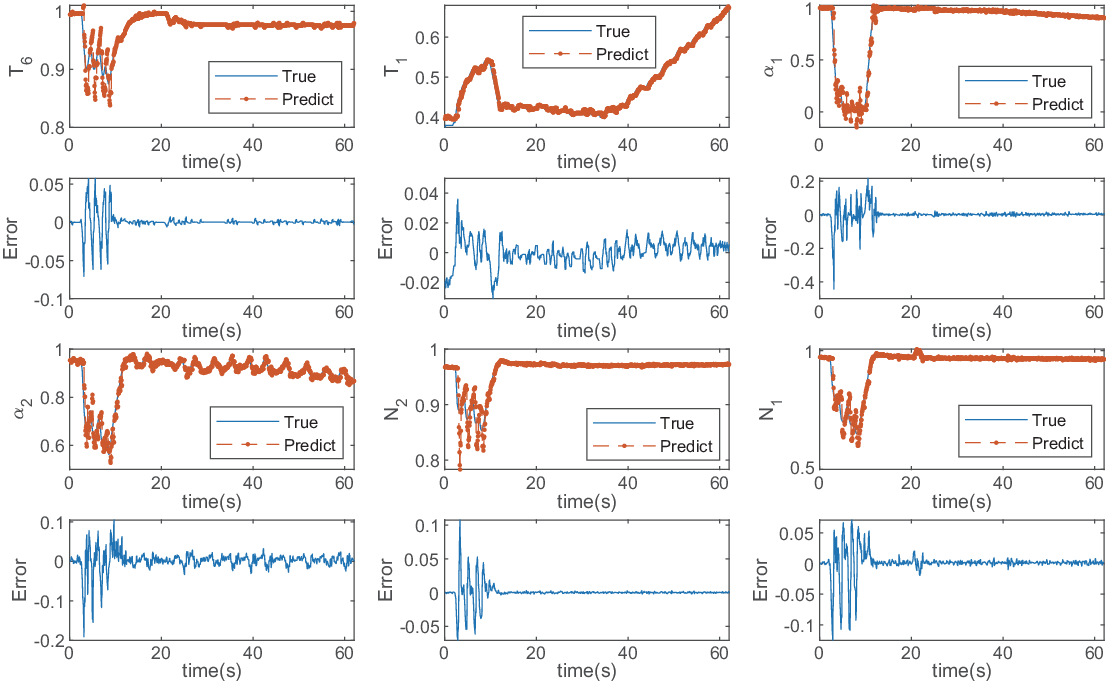}
	\caption{{Prediction results of aero-engine parameters $T_6$, $T_1$, $\alpha_{1}$, $\alpha_{2}$, $N_2$, $N_1$ when the prediction step is 10.}}\label{fig7}
\end{figure*}

\noindent When the time step is 1, the prediction results of the aero-engine parameters by Spike-ESN model are shown in Figure \ref{fig6}, where the fault occurs in the range of 1s to 15s. It can be seen from Figure \ref{fig6} that the prediction errors of the Spike-ESN model are about ${{10}^{-4}}$, which can accurately reflect the development trend of aero-engine state data in the short term. When the prediction time step increases to 10, the prediction results of the aero-engine parameters by Spike-ESN model are shown in Figure \ref{fig7}, where almost all of the prediction errors are about ${{10}^{-2}}$. Although the prediction error of the Spike-ESN model increases with the increase of time step, it can still predict the development trend of aero-engine states in the future. 

\begin{figure*}[ht]
	\centering
	\includegraphics[height=0.5\textwidth,width=1.\hsize]{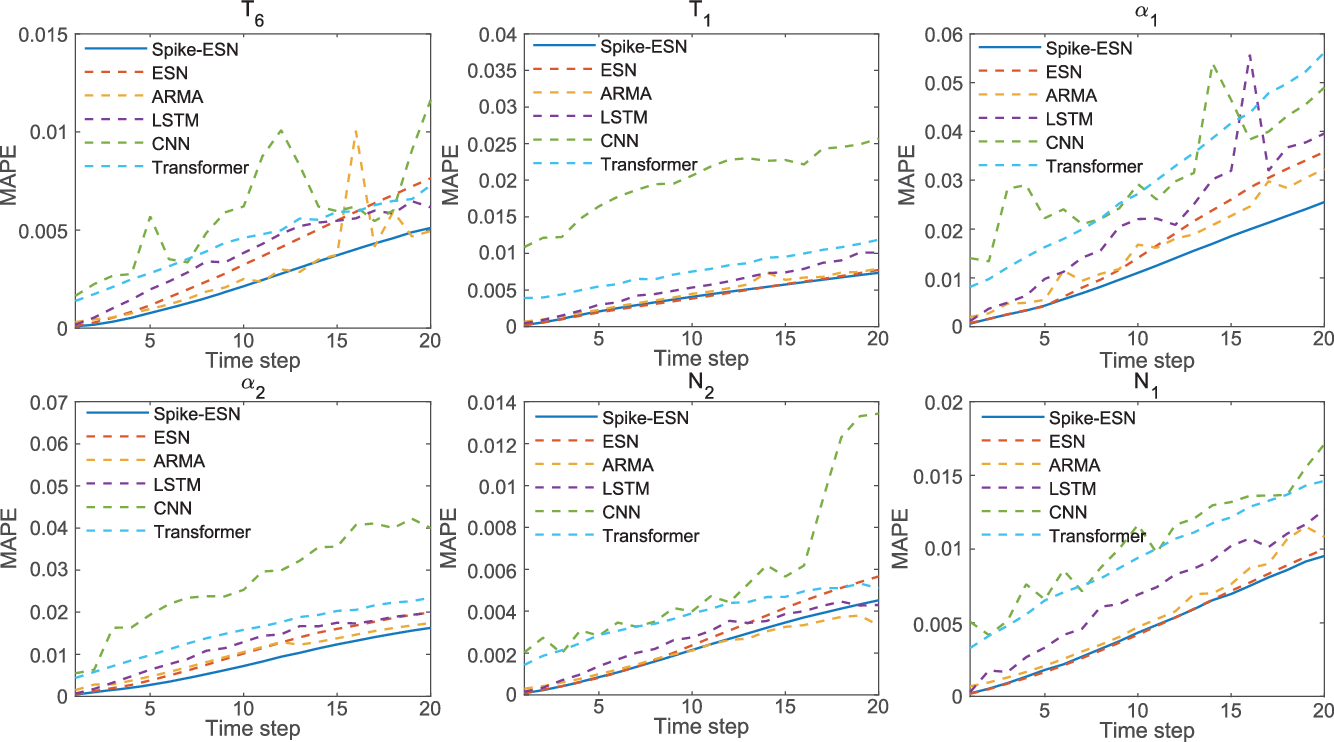}
	\caption{{MAPE comparison of various aero-engine state parameters in different prediction steps of six models.}}\label{fig9}
\end{figure*}

Specifically, the six aero-engine state parameters are predicted by six models when the time step varies from 1 to 20 (i.e., as shown in Figure \ref{fig9}), and the corresponding practical physical significance is to predict the changes of aero-engine parameters in the next 0.062s to 1.24s. The prediction errors of the six models with time step of 1, 10 and 20 for each parameter are sorted as shown in Table \ref{tab1}.

From Figure \ref{fig9}, it can be noticed that the Spike-ESN model is superior to the ESN model in MAPE. Moreover, with the increase of prediction time step, the convergence of Spike-ESN is better than that of ESN. It can be seen from Table \ref{tab1} that Spike-ESN can achieve better performance than ESN in the prediction of all parameters in the experiment. Especially, the accuracy of all parameters has been improved by $2\text{\textperthousand}$ in the long-term prediction with a step size of 20, which may play a key role in aero-engine parameter prediction and fault diagnosis.
In a short prediction step, the error of Spike-ESN model is ${{10}^{-3}}$ less than that of ESN model and ARMA model, and with the increase of prediction time step, the advantage of the Spike-ESN model is gradually obvious.

As a traditional mathematical calculation method, ARMA can obtain better results in some prediction steps and has fast training speed, but its convergence and stability are generally worse than those of the other neural networks. ARMA may get very poor results in some prediction steps, so its error curve is not smooth.

In comparison, the RMSE and MAPE of the CNN are almost largest, and the error fluctuation of the CNN is wide. The experimental results show that the CNN possesses a poor ability to extract time series features. The reason is that there is no memory unit with recurrent serial structure in the CNN. Therefore, it is difficult for CNNs to combine a large amount of past time information for analysis.

As shown in Figure \ref{fig9}, the error of LSTM increases with the increase of the prediction step, but always at a low level. As the prediction step increases, LSTM gradually shows the effect of exceeding other models, which is due to the advantage of LSTM's recurrent memory unit for time series information. Comparing the running time, the single run time of LSTM is 99.56s, which is much faster than that of CNN.

From Figure \ref{fig9}, the prediction effect of Transformer is very close to that of CNN, which is caused that both CNN and Transformer are parallel input-output structures without memory units. Transformer relies on positional encoding to capture time-series information which is less effective in time series. In addition, Transformer is weak in predicting non-stationary time series since it lacks the ability to adapt well to changes in time series data. Further, Transformer's multi-headed attention mechanism is more suitable for large-scale data, such that it is difficult to achieve better results in the small-scale data in this paper. Moreover, the Transformer has a single run time of 155.62s which can also only be used for offline aero-engine fault diagnosis on the ground.

\begin{table*}[h]
	\centering
	\renewcommand{\arraystretch}{1.2}
	\caption{{Main differences between the 6 networks (CNN, Transformer, LSTM, ARMA, ESN, Spike-ESN).}}\label{tab3}
	\begin{tabular}{cccccccc}
		\hline
		\multirow{2}{*}{Model}&\multicolumn{7}{c}{Main Differences}\\
		&{Bio-interpretability}&{Spatial-temporal Dynamics}&{Memory}&{Sparsity}&{Stability}&{Training Speed}&{Accuracy}\\
		\hline
		CNN&No&No&No&No&Low&Low&Low\\
		Transformer&No&{No}&No&No&{Medium}&Low&Medium\\
		{LSTM}&No&No&{Yes}&No&{Medium}&{Low}&{High}\\
		{ARMA}&No&No&No&No&Low&{Medium}&{High}\\
		{ESN}&No&No&{Yes}&Yes&{High}&{High}&{High}\\
		{\textbf{Spike-ESN}}&\textbf{Yes}&\textbf{{Yes}}&\textbf{{Yes}}&\textbf{{Yes}}&\textbf{{High}}&\textbf{{High}}&\textbf{{Best}}\\
		\hline
	\end{tabular}
\end{table*}
It is important to note that our dataset is a professional dataset with complex properties from aero-engine research institute. The properties of the datasets have a great impact on the prediction accuracy, so it is one-sided to judge the model's performance only from the prediction accuracy. If the predicted time series data are smooth and stable, the RMSE of the poorer prediction results will also exhibit small. In this case, a model with weak predictive ability will exactly match the target if it holds the value of the previous moment. On the contrary, the data of parameter $\alpha_{2}$ in the dataset after the failure are mostly fluctuating, so it is a greater challenge for the prediction algorithm, which can distinguish the performance of the model prediction. For instance, other deep networks such as the LSTM, which are able to extract deep features of time series and have a large number of training parameters, have an advantage in long-time prediction capability. In contrast, our method has the advantage of small parameter size and fast training speed, for which a comparison of the single training time of the model is added. Comparing single run time, Spike-ESN takes only 8.31s to complete the training at a spike sampling count of 50, and only 4.92s to complete the training at a spike sampling count of 20. But all other deep large-scale networks have difficulty to achieve such speed.

As noted above, the data can be spiked by using Spike-ESN model, which can be reflected in the spike space. The Spike-ESN model converts sequence data into spatio-temporal information, which is more conducive for the model to extracting time factors. Then, the temporal and spatial information is converted into the external input current of the ESN neuron by using the spike conversion formula, so that the Spike-ESN model can capture sufficient features. The reason why Spike-ESN is a more stable and accurate model than ESN and ARMA is that Spike-ESN model includes the advantages of small amount of ESN calculation, convenient training and being suitable for time series, meanwhile, the spike mechanism is added to make the new model more sensitive to the time information in the time series. In long-term prediction, it is difficult for the model to establish the input-output mapping relationship for complex signals. 
Therefore, a higher requirement is put forward for the spatio-temporal feature extraction of the Spike-ESN model. The experimental results show that Spike-ESN has a stronger long-term prediction ability than ESN, and can eliminate the error accumulation of long-term prediction.
Moreover, Spike-ESN does not require iterative optimization, so the training speed is quite fast. Compared with several networks above, Spike-ESN has the characteristics of high prediction accuracy and fast training speed, which is better at small-scale time series prediction.

Based on the above analysis, Table \ref{tab3} is presented to list the differences between the six models (CNN, Transformer, LSTM, ARMA, ESN and Spike-ESN) from seven aspects (Bio-interpretability, Spatial-temporal Dynamics, Memory, Sparsity, Stability, Training Speed and Accuracy). As shown in Table \ref{tab3}, Spike-ESN has unique bio-interpretability, spatio-temporal dynamics and memory to extract time series features. Besides, it has high performance in terms of stability and accuracy. Due to the sparse structure and the special training method, the training speed of proposed method is high.

\begin{table*}[ht]
	\centering
	\caption{{Prediction errors (RMSE and MAPE) for predicting 6 parameters ($T_6$, $T_1$, $N_2$, $N_1$, $\alpha_1$, $\alpha_2$) by 6 models (CNN, Transformer, LSTM, ARMA, ESN, Spike-ESN) at prediction step of 1, 10, 20.}}\label{tab1}
	\begin{tabular}{ccccccccccccccc}
		\hline
		\multirow{2}*{Prediction Step}&\multirow{2}*{Model}&\multicolumn{6}{c}{RMSE($\times {{10}^{-3}}$)}&&\multicolumn{6}{c}{MAPE($\times {{10}^{-3}}$)}\\
		&&{$T_6$}&{$T_1$}&{$N_2$}&{$N_1$}&{$\alpha_1$}&{$\alpha_2$}&&{$T_6$}&{$T_1$}&{$N_2$}&{$N_1$}&{$\alpha_1$}&{$\alpha_2$}\\
		\hline
		\multirow{6}*{Step-1}&CNN&5.62&21.3&6.08&11.7&26.6&14.1&&1.65&10.9&2.03&5.06&14.0&5.49\\
		&Transformer&4.68&6.63&4.64&9.49&27.1&12.9&&1.39&3.90&1.43&3.29&8.15&4.38\\
		&LSTM&1.85&1.42&0.903&1.26&5.90&2.86&&0.181&0.469&0.156&0.308&1.18&0.683\\
		&ARMA&1.51&1.66&1.63&3.08&10.3&5.27&&0.313&0.683&0.299&0.696&1.99&1.50\\
		&ESN&0.532&0.599&0.645&\textbf{0.933}&\textbf{2.65}&1.82&&0.106&0.224&0.0968&\textbf{0.190}&0.768&0.549\\
		&\textbf{Spike-ESN}&\textbf{0.503}&\textbf{0.588}&\textbf{0.551}&0.945&2.81&\textbf{1.61}&&\textbf{0.0936}&\textbf{0.217}&\textbf{0.0851}&0.209&\textbf{0.632}&\textbf{0.426}\\
		\hline
		\multirow{6}*{Step-10}&CNN&18.6&38.9&15.6&32.9&80.8&54.7&&6.20&20.6&3.99&11.7&29.2&25.3\\
		&Transformer&15.5&14.8&14.9&32.6&106&46.5&&4.60&7.53&3.90&9.41&27.2&15.8\\
		&LSTM&12.4&11.0&\textbf{10.8}&24.1&81.0&38.2&&3.84&5.35&2.75&6.91&22.0&12.9\\
		&ARMA&\textbf{10.6}&10.0&10.9&19.8&73.4&31.8&&2.51&4.49&\textbf{2.11}&4.72&16.8&10.5\\
		&ESN&11.9&8.25&13.1&19.6&42.1&28.9&&3.23&\textbf{3.84}&2.37&\textbf{4.19}&14.1&10.2\\
		&\textbf{Spike-ESN}&10.7&\textbf{8.07}&12.2&\textbf{18.9}&\textbf{40.7}&\textbf{23.1}&&\textbf{2.14}&4.08&2.15&4.33&\textbf{11.0}&\textbf{7.23}\\
		\hline
		\multirow{6}*{Step-20}&CNN&25.8&45.8&30.6&50.4&157&84.0&&11.6&25.7&13.4&17.1&49.0&40.0\\ 
		&Transformer&21.2&23.9&20.1&50.4&179&66.0&&7.28&11.9&5.09&14.6&56.2&23.5\\
		&LSTM&\textbf{18.9}&20.1&\textbf{16.3}&43.0&146&60.4&&6.16&10.1&4.30&12.7&39.6&19.8\\
		&ARMA&22.8&17.2&\textbf{16.3}&38.0&106&50.2&&\textbf{4.94}&7.93&\textbf{3.33}&10.8&32.3&17.4\\
		&ESN&28.9&17.2&34.8&39.9&101&53.6&&7.64&7.74&5.66&10.0&35.8&19.9\\
		&\textbf{Spike-ESN}&24.5&\textbf{14.7}&24.6&\textbf{37.6}&\textbf{99.6}&\textbf{47.1}&&5.11&\textbf{7.35}&4.52&\textbf{9.54}&\textbf{25.5}&\textbf{16.3}\\
		\hline
	\end{tabular}
\end{table*}
\begin{figure}[ht]
	\centering
	\includegraphics[height=0.4\textwidth,width=1.\hsize]{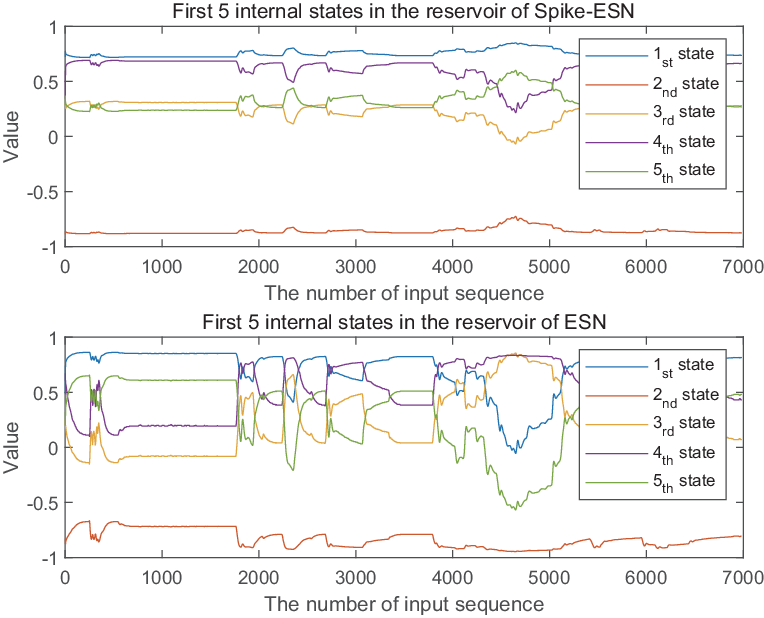}
	\caption{{The first 5 internal states of the reservoirs in Spike-ESN and ESN for predicting $\alpha_2$ at step of 20.}}\label{fig10}
\end{figure}
\begin{figure}[ht]
	\centering
	\includegraphics[height=0.4\textwidth,width=1.\hsize]{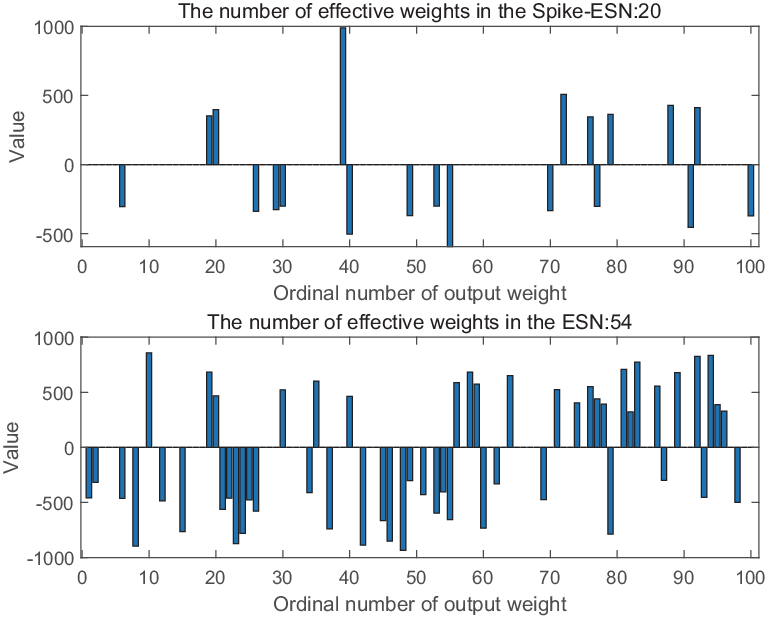}
	\caption{{The output weight matrices in Spike-ESN and ESN for predicting $\alpha_2$ at step of 20.}}\label{fig14}
\end{figure}

To sum up, the Spike-ESN model has enhanced time information feature extraction, which can better learn the features of state instability and rotating stall process before aero-engine surge fault. Furthermore, it can realize the role of short-term early prediction parameters in aero-engine fault diagnosis, which provides favorable conditions for aero-engine to prevent faults in the early stage of surge. the Spike-ESN model has achieved good results on the whole, which reflects the data that will occur in the future of aero-engines more accurately. In this experiment, Spike-ESN model can predict engine parameters more accurately and stably after 1.24s than ESN and ARMA. Significantly, without being affected by sudden changes in the external environment or personnel operation, users can roughly know the possible abnormalities of aero-engine parameters after 1.24s, and then make rapid response.  If the aircraft is suddenly affected, the rapid training of the Spike-ESN model can update the latest prediction in real time.

\subsection{Discussion}
\noindent In this subsection, we will discuss the benefits of spike encoding and spatial projection, and experimentally explore the effect of spike encoding length on the performance.

\begin{figure}[t]
	\centering
	\includegraphics[height=0.4\textwidth,width=1.\hsize]{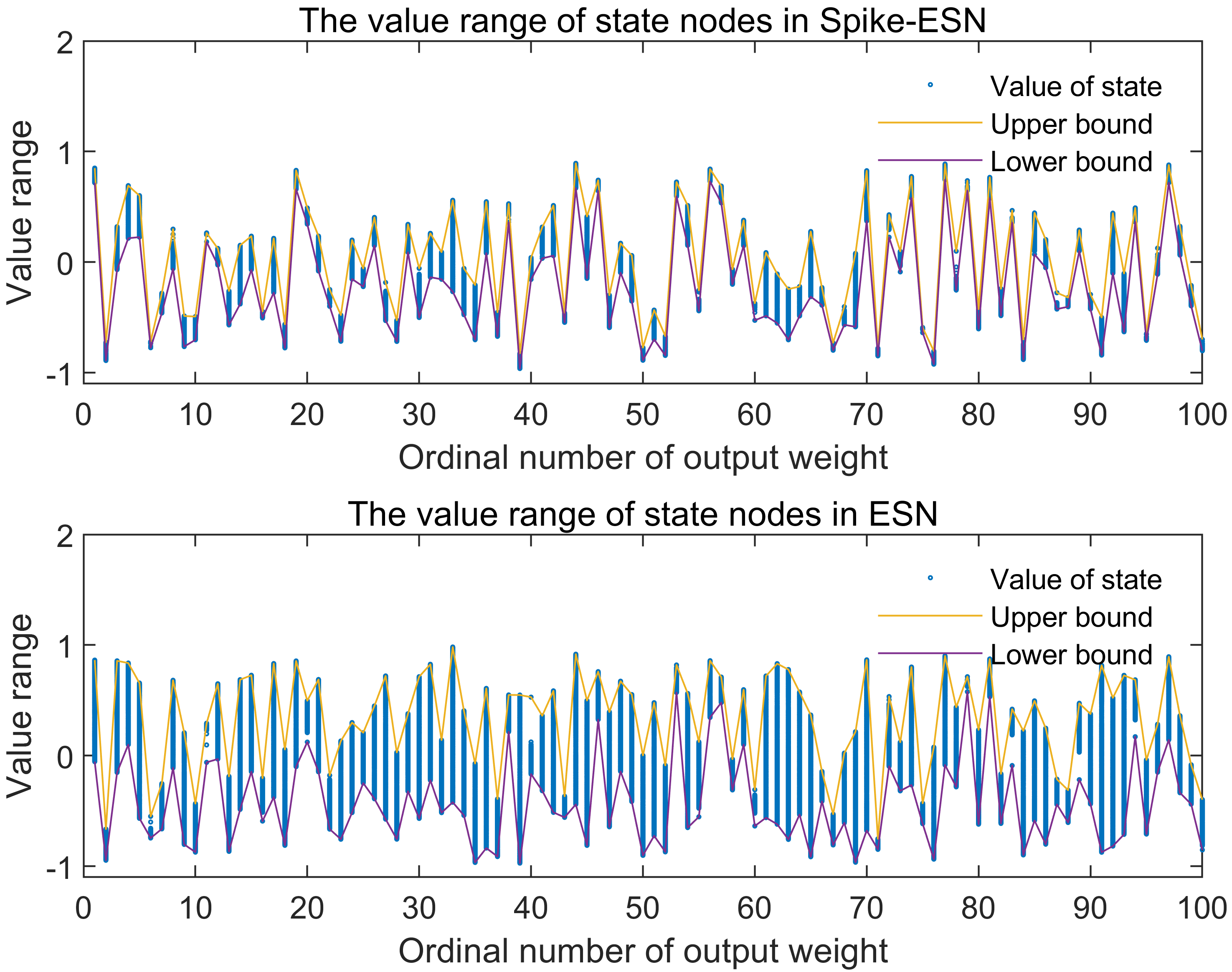}
	\caption{Value range of the states in the reservoirs of Spike-ESN and ESN for predicting $\alpha_2$ at step of 20.}\label{fig15}
\end{figure}

1) In order to more fully demonstrate the prediction performance of the proposed method, it is necessary to visualize the deep features of the network. Since the internal state of the reservoir can clearly reflect the deep features in the Spike-ESN, the process of reading out the internal state needs to be visualized. As shown in (\ref{EQ9}), it is clear to reflect the prediction effect by visualizing internal state matrix $\textbf{X}$ and output weight $\textbf{W}_{out}$ in Figures \ref{fig10} and \ref{fig14}, respectively.

The five curves in Figure \ref{fig10} represent the time series of the five neuron state in the internal state matrix $\textbf{X}$ when updated with the input data. Hence, the x-axis in Figure \ref{fig10} represents the number of input sequence data and the y-axis represents the value of the internal state. In order to compare the effects of the proposed spike encoding method on the deeper features of the network, the first and second subplots in Figure \ref{fig10} visualize the internal states of Spike-ESN and ESN, respectively. As shown in Figure \ref{fig10}, after spike encoding and neuronal membrane currents for temporal feature extraction, the curves in the first subplot cross each other less, while almost all the curves in the second subplot cross each other a lot, i.e., the input features make the internal states of the network separated and smooth as far as possible.

To explain the benefits of separating and smoothing the state variables, the value range of each state node is plotted in Figure \ref{fig15}. In terms of value range, separating and smoothing increase the value range of the main state nodes and reduce that of the redundant nodes. As shown in Figure \ref{fig15}, the width of the value range between the state nodes in Spike-ESN is significantly different, while the value range of the state nodes in ESN is almost the same. As shown in (\ref{EQ9}), a linear regression method was used to read out the internal state, which can be written as:
\begin{eqnarray}\label{EQ12}
	\hat y = {w_1}{x_1} + {w_2}{x_2} +  \cdots  + {w_{{N_{res}}}}{x_{{N_{res}}}},
\end{eqnarray}
where, for $i\in \left\{ 1,2,\cdots {{N}_{res}} \right\}$, $x_i$ represents the internal state, $w_i$ represents the corresponding weight coefficient of each internal state.

The width of the value range of the state node is denoted as $\delta {x_i}$, and the width of the value range of the predicted result is denoted as $\delta y$. Then, for each input data corresponding to ${x_i}$, the value range $\delta {x_i}$ contains both beneficial and harmful components for accurate prediction. If the scale of $\delta {x_i}$ is small, then the corresponding ${x_i}$ is barely changed when the input data is changed. Its contribution to all predictions varies so little that it hardly affects the prediction results. On the contrary, if the scale of $\delta {x_i}$ is large, the corresponding ${x_i}$ changes a lot when the input data is changed, which has a large impact on the prediction results. Further, the state node contains sufficient features, i.e., the beneficial component is large and the harmful component is small, then the prediction results will be more accurate. From Table \ref{tab1}, it can be seen that the RMSE of Spike-ESN is smaller than that of ESN, i.e., the prediction accuracy is better.

Correspondingly, the first and second subplots in Figure \ref{fig14} visualize the output weights $\textbf{W}_{out}$ for Spike-ESN and ESN, respectively. In Figure \ref{fig14}, the x-axis and the y-axis indicate ordinal number and value of weights, respectively. For example, the number $25$ on the x-axis indicates the $25_{th}$ weight in the output weights. From the comparison experiments, it is found that several output weights in Spike-ESN exhibit large values, while most output weights have low values. Since the output weights correspond to the internal states, it can be seen that Spike-ESN concentrates the redundant components in some internal states, which facilitates to reduce their impact on the prediction effect. On the contrary, the output weights of ESN are uniformly distributed, which will enhance the effect of the redundant components caused during the high-dimensional projection of the features on the output results. In order to clearly distinguish the difference between the output weights $\textbf{W}_{out}$ in the comparison experiments, a threshold is set to filter out the output weights that significantly affect the network performance. In Figure \ref{fig14}, when the threshold is set to 300, there are 20 and 54 significant output weights for Spike-ESN and ESN, respectively. In addition, when the threshold is set to 200, there are 36 and 70 significant output weights for Spike-ESN and ESN, respectively. When the threshold is set to 400, there are 7 and 46 significant output weights for Spike-ESN and ESN, respectively. Since multiple sets of data illustrate the same law, only one case in Figure \ref{fig14} is shown, and other settings are omitted.

From Figure \ref{fig14}, it is demonstrated that these 20 weights reflect the time series information more accurately. Further, the value of the beneficial component is large in state variables with large value range width $\delta {x_i}$. Correspondingly, there are 54 effective weights for ESN, which proved that the information of the time series is not accurately reflected in these states. Meanwhile, the value of the harmful component is large in the states with large value range width $\delta {x_i}$. Therefore, it can be concluded that spike encoding has a certain enhancement effect on the extraction of state features. To sum up, the states in the first subplot of Figure \ref{fig10} are more suitable as a base for the linear regression.

Finally, since a large prediction step can better reflect the advantages of the spike encoding approach in temporal feature extraction, we use the sequence with a prediction step of 20 to compare the internal states. The first and second subplots of Figure \ref{fig10} and Figure \ref{fig14} represent the internal states and the corresponding output weights of Spike-ESN and ESN for predicting $\alpha_2$ at the prediction step of 20, respectively. From Table \ref{tab1}, the prediction error RMSE of Spike-ESN is 0.0471 and that of ESN is 0.0536. It can be seen that the proposed method has a significant advantage, which is consistent with the above analysis.

\begin{figure}[t]
	\centering
	\includegraphics[height=0.4\textwidth,width=1.0\hsize]{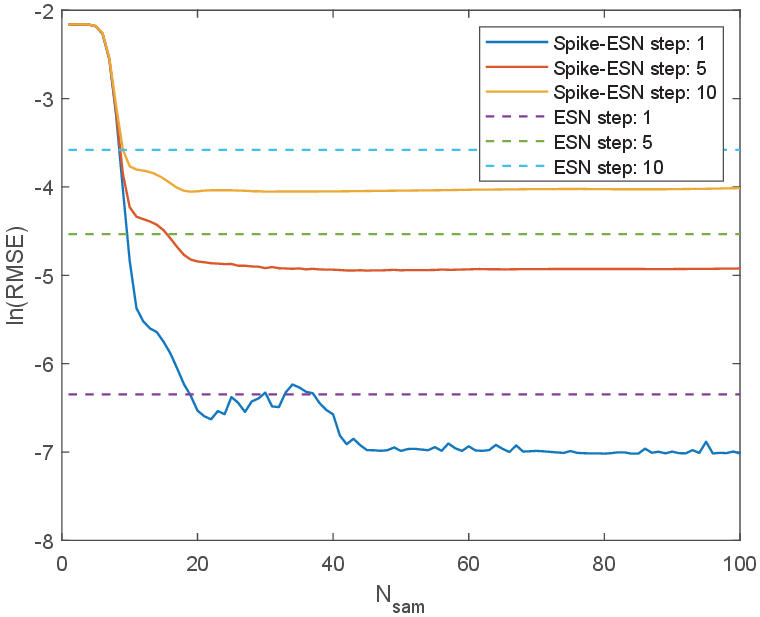}
	\caption{{The influence of spike sampling times on model accuracy.}}\label{fig12}
\end{figure}

2) Traditionally, the echo state network maps the input signal into a sparse high-dimensional space by a random nonlinear mapping method \cite{jaeger2001echo}. Further, the input signal has good spatial features, so that a simple linear method for the output weight training can make the model achieve excellent performance. Therefore, the rationality of this high-dimensional projection directly affects the quality of the optimization results. However, the echo state network does not have the spatio-temporal dynamics similar to biological neurons, which make it hard to capture the temporal features of time series data. In reality, neurons in biological systems use spikes to capture and transmit signals, which cause neurons to generate membrane potential firing. Typically, we can approximate neuron dynamics as an integral process. In particular, when the neuron membrane potential is higher than a certain threshold voltage, the neuron will emit an action potential to transmit information \cite{subbulakshmi2021biomimetic}. In order to transmit different information through action potentials, the size and firing time of the neuron's action potential are critical.

In conclusion, the spiking neuron model can map a physical continuous information flow into a discrete dynamic system in a high-dimensional neural information space, which greatly improves feature extraction and data analysis. Similarly, the high-dimensional projection of spatio-temporal dynamics is more beneficial to the output layer of the echo state network to extract the necessary features. 

To verify the effect of spatio-temporal dynamics in Spike-ESN, the test data are collected from actual aero-engine sensors. We adjust the spike sampling frequency to vary the input current sequence length of spike neurons, and then test a signal prediction effect at several time steps separately. Meanwhile, an ESN model without temporal dynamics was used for comparative testing, where the two models share the same key parameters. Since the scales of test results vary greatly, the results are represented by the natural logarithm (i.e., as shown in Figure \ref{fig12}). The parameter $N_{sam}$ is the sampling times of the spiking neuron, i.e., the dimension of the temporal projection. After the triggering of the spike sequence, the membrane potential of the spiking neuron generates a current sequence with the time length of $N_{sam}$. Then the current sequence is used as the input of the network, which makes the input data projected into a high-dimensional space in time for feature extraction. Figure \ref{fig12} shows the prediction results on the dataset by adjusting the value of $N_{sam}$ from 1 to 100. Thus, the effect of $N_{sam}$ (i.e., the temporal projection dimension) on the network performance is obtained. It can be seen from Figure \ref{fig12} that the higher the temporal projection dimension, the stronger the performance of the network, reflecting the better feature extraction ability. However, after the time dimension is increased to a certain level, the network performance no longer increases, which mean that the features extracted by the projection method are exhausted.

The test results show that the RMSE of Spike-ESN has a significant decrease with the increase of the number of spike sampling, which exceeds the performance of ESN at each prediction step. Based on these results, the introduction of spatio-temporal dynamics and increasing the spike sampling frequency have a positive effect on the prediction results of the model.

 \section{Concluding remarks}
\noindent In this work, we designed a spike echo state network (Spike-ESN) model based on the deep learning framework to solve the problem of aero-engine future state prediction. First, the spike input layer based on Poisson distribution in Spike-ESN model can extract useful temporal features from aero-engine sequence data. Then, Spike-ESN model inputs the time characteristics into the spike reservoir through the spike accumulation calculation method, which can project the data into high-dimensional sparse space. Spike-ESN model adopts spike neurodynamics mechanism and time sensitive echo state mechanism, which has the ability to enhance the temporal information characteristics of data and time series memory. Based on the results of aero-engine future state prediction, it is proved that Spike-ESN model is  a method with high accuracy and stability. Spike-ESN model can effectively predict the future flight parameters of aero-engines, which lays a good foundation for aero-engine early fault warning.



\bibliographystyle{unsrt}       
\bibliography{mybibfile}
\begin{IEEEbiography}[{\includegraphics[width=1in,height=1.25in,clip,keepaspectratio]{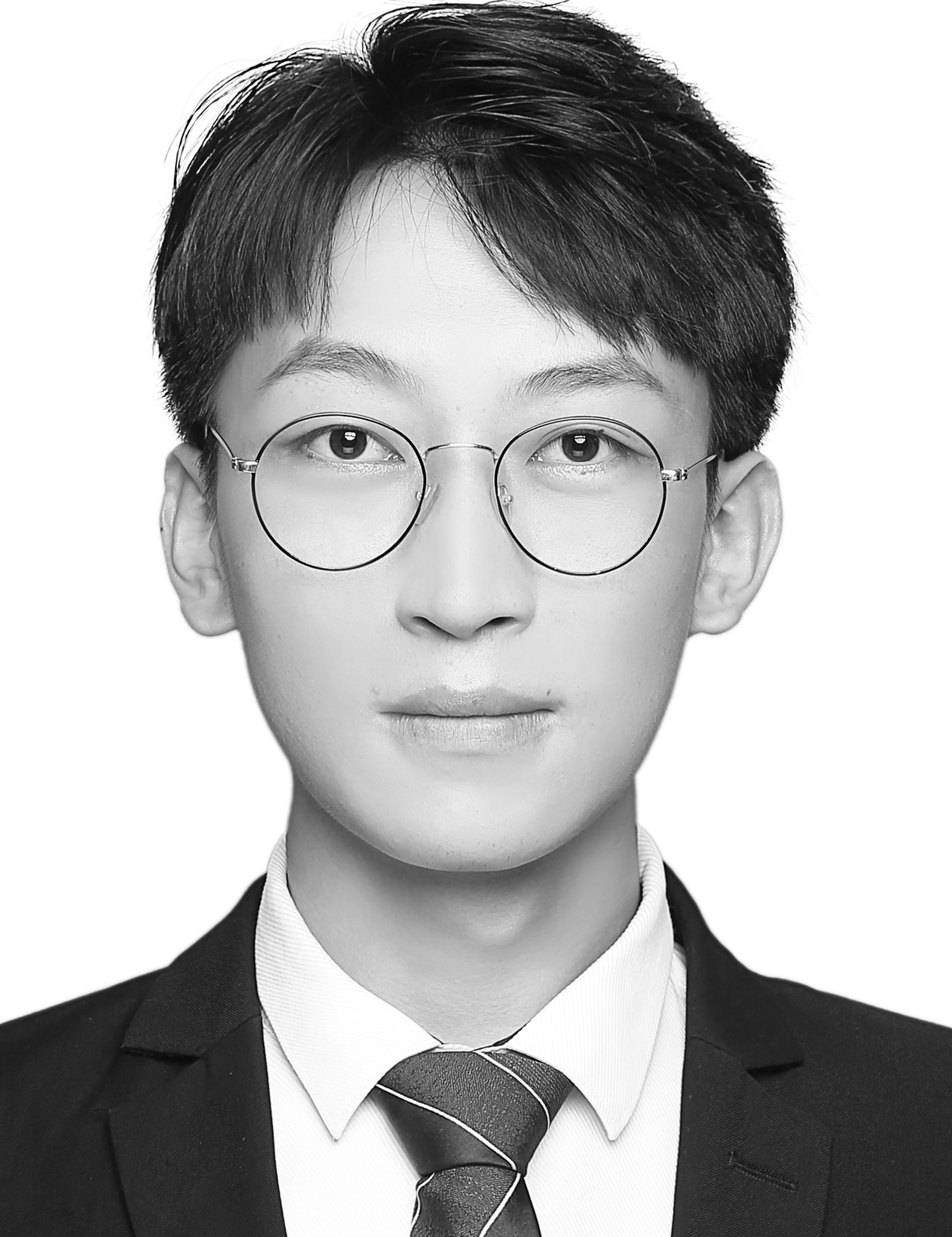}}]{Mo-Ran Liu}
	received the B.S. degree in control theory and control engineering with the School of Control Science and Engineering, Dalian University of Technology, Dalian, China, in 2022, where he is currently pursuing the M.S. degree in control theory and control engineering with the School of Control Science and Engineering. His research interests include brain-inspired intelligence, fault diagnosis, aero-engine control, and so on.\end{IEEEbiography}

\begin{IEEEbiography}[{\includegraphics[width=1in,height=1.25in,clip,keepaspectratio]{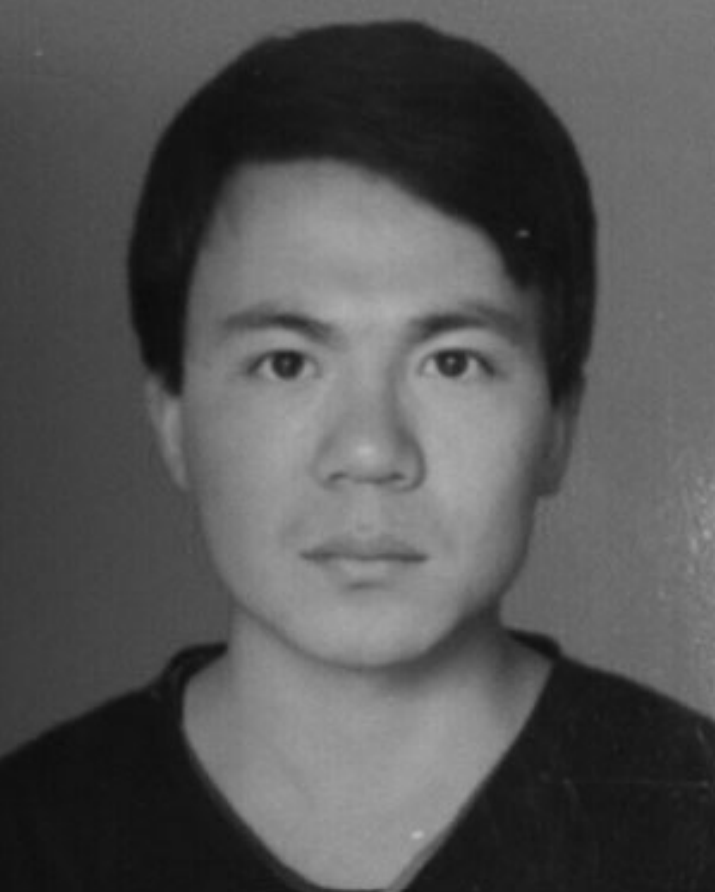}}]{Tao Sun}
received the M.S. degree in Operational
Research and Cybernetics from the School of Mathematical Sciences, Dalian University of Technology,
Dalian, China, in 2017, and the Ph.D. degree in
Control Theory and Control Engineering from the
School of Control Science and Engineering, Dalian
University of Technology, Dalian, China, in 2022.
He is currently a Post-Doctoral Researcher with
the Department of Automation, Tsinghua University.
His research interests include brain-inspired intelligence, fault diagnosis, aero-engine control, and so on.
Mr. Sun serves as a Reviewer for the Mathematical Reviews in the
American Mathematical Society and an Editorial Board Member for the
Journal of Modern Industry and Manufacturing, International Journal of
Systems Engineering, International Journal of Data Science and Analysis,
and Machine Learning Research.\end{IEEEbiography}

\begin{IEEEbiography}[{\includegraphics[width=1in,height=1.25in,clip,keepaspectratio]{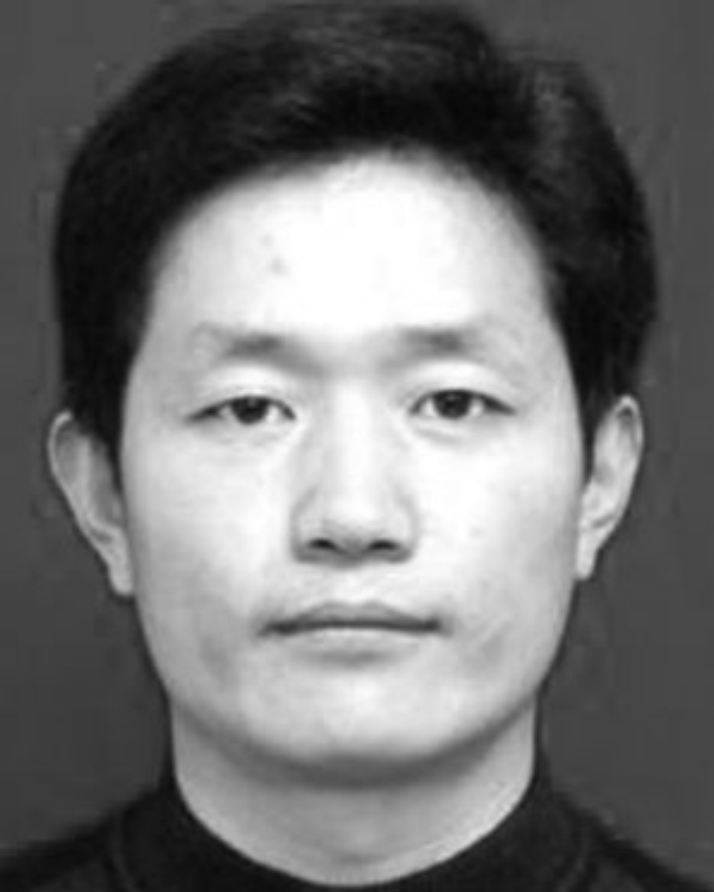}}]{Xi-Ming Sun (Senior Member, IEEE)}
	received the
	M.S. degree in applied mathematics from Bohai
	University, Jinzhou, China, in 2003, and the Ph.D.
	degree in control theory and control engineering
	from Northeastern University, Shenyang, China, in
	2006.
	From 2006 to 2008, he was a Research Fellow
	of the Faculty of Advanced Technology, University
	of Glamorgan, Pontypridd, U.K. He then visited
	the School of Electrical and Electronic Engineering,
	Melbourne University, Melbourne, VIC, Australia,
	in 2009, and the Polytechnic Institute of New York University, New York,
	NY, USA, in 2011. He is currently a Professor in the subject of control theory
	and control engineering with the School of Control Science and Engineering,
	Dalian University of Technology, Dalian, China. His research interests include
	artificial intelligence control, network control, and aero-engine control.
	Prof. Sun was a recipient of the Most Cited Article 2006–2010 Award from
	the journal Automatica in 2011. He serves as an Associate Editor for the IEEE
	TRANSACTIONS ON CYBERNETICS and Chinese Journal of Aeronautics.\end{IEEEbiography}
\end{document}